\documentclass[11pt,a4paper]{article}
\usepackage[hyperref]{acl2018}
\usepackage{times}
\usepackage{latexsym}

\usepackage{url}

\usepackage{graphicx}  
\usepackage{amsmath}
\usepackage{algorithm}
\usepackage{algpseudocode}
\usepackage{amsfonts}
\usepackage{amssymb}
\usepackage{bm}
\usepackage{color}
\usepackage{pgfplots}
\usepackage{subcaption}

\newcommand{\MemNet}{\mathbf{\textrm{MemNet}}}
\newcommand{\vE}{{\bm{E}}}
\newcommand{\va}{{\bm{a}}}
\newcommand{\vs}{{\bm{s}}}
\newcommand{\vc}{{\bm{c}}}
\newcommand{\vv}{{\bm{v}}}
\newcommand{\vh}{{\bm{h}}}
\newcommand{\vd}{{\bm{d}}}
\newcommand{\vx}{{\bm{x}}}
\newcommand{\vm}{{\bm{m}}}
\newcommand{\vy}{{\bm{y}}}
\newcommand{\vb}{{\bm{b}}}
\newcommand{\vq}{{\bm{q}}}

\newcommand{\vp}{{\bm{p}}}
\newcommand{\vr}{{\bm{r}}}
\newcommand{\vW}{{\bm{W}}}
\newcommand{\vtheta}{{\bm{\theta}}}
\newcommand{\valpha}{{\bm{\alpha}}}
\newcommand{\vM}{{\bm{M}}}

\newcommand{\RNN}{\textrm{RNN}}
\newcommand{\softmax}{\textrm{softmax}}
\newcommand{\unk}{\texttt{<UNK>} }
\newcommand{\srcmem}{\textit{S-NMT+src mem}}
\newcommand{\trgmem}{\textit{S-NMT+trg mem}}
\newcommand{\bothmem}{\textit{S-NMT+both mems}}
\newcommand{\snmt}{\textit{S-NMT}}

\newcommand{\reza}[1]{{\color{black} #1}} 
\newcommand{\sameen}[1]{{\color{black} #1}} 

\aclfinalcopy 

\title{Document Context Neural Machine Translation with Memory Networks}

\author{Sameen Maruf \and Gholamreza Haffari\\
  Faculty of Information Technology, Monash University, Australia\\
  {\tt \{firstname.lastname\}@monash.edu} \\}

\date{}

\begin{document}
\maketitle
\begin{abstract}
 We present a document-level neural machine translation model which takes both source and target document context into account using memory networks.
We model the problem as a structured prediction problem with interdependencies among the observed and hidden variables, i.e., the source sentences and their unobserved target translations in the document. 
The resulting structured prediction problem is tackled with a neural translation model equipped with two  memory components, one each for the source and target side, to capture the documental interdependencies.
We train the model end-to-end, and propose an iterative decoding algorithm based on block coordinate descent.
Experimental results of English translations from French, German, and Estonian documents show that our model is effective in exploiting both source and target document context, and statistically significantly outperforms the previous work in terms of BLEU and METEOR.  
\end{abstract}

\section{Introduction}
Neural machine translation (NMT) has proven to be powerful \cite{Sutskever:14,Bahdanau:15}. It is on-par, and in some cases, even surpasses the traditional statistical MT \cite{luong-pham-manning:2015:EMNLP}  while enjoying more flexibility and significantly less manual effort for feature engineering. Despite their flexibility, most neural MT models  translate sentences independently. 
%
%
{Discourse phenomenon such as pronominal anaphora and lexical consistency, may depend on long-range dependency going farther than a few previous sentences, are neglected in sentence-based translation \cite{Bawden:17}.}
%

There are only a handful of attempts to document-wide machine translation in statistical and neural MT camps. 
\newcite{Hardmeier:10,Gong:11,Garcia:14} propose document translation models based on statistical MT \sameen{but are restrictive in the way they incorporate the document-level information and fail to gain significant improvements.} 
More recently, there have been a few attempts to incorporate source side context into neural MT \cite{Jean:17,Wang:17, Bawden:17};  however, these works only consider a very local context including a few previous source/target sentences, ignoring the global source and target documental contexts. {The latter two report deteriorated performance when using the target-side context.}

In this paper, we present a document-level machine translation model which combines sentence-based NMT  \cite{Bahdanau:15} with memory networks \cite{Sukhbaatar:15}. We capture the global source and target document context with two memory components, one each for the source and target side, and incorporate it into the sentence-based NMT by changing the decoder to condition on it as the sentence translation is generated.
%
%
\reza{We conduct experiments on three language pairs: French-English, German-English and Estonian-English. 
The experimental results and analysis demonstrate that our model is effective in exploiting both source and target document context, and statistically significantly outperforms the previous work in terms of BLEU and METEOR.
}

\section{Background}
\subsection{Neural Machine Translation (NMT)}
Our document NMT model is grounded on sentence-based NMT model \cite{Bahdanau:15} which contains an encoder to \emph{read} the source sentence as well as an attentional decoder to \emph{generate} the target translation.
 

\paragraph{Encoder}
%
%
It is a bidirectional RNN consisting of two RNNs running in opposite directions
over the source sentence:

\vspace{-3mm}
{\small
$$\overrightarrow{\vh_{i}}= \overrightarrow{\RNN}( \overrightarrow{\vh}_{i-1}, \vE_S [x_{i}]),
\overleftarrow{\vh}_{i} = \overleftarrow{\RNN}(\overleftarrow{\vh}_{i+1}, \vE_S [x_{i}]) 
$$
}
where $\vE_S[x_i]$ is embedding of the word $x_i$ from the embedding table $\vE_S$ of the source language,  and $\overrightarrow{\vh}_{i}$ and $\overleftarrow{\vh}_{i}$ are the hidden states of the forward and backward RNNs which can be based on the LSTM  \cite{Hochreiter:97} or GRU  \cite{cho_gru2014} units. Each word in the source sentence is then represented by the concatenation of the corresponding bidirectional hidden states, $\vh_i=[\overrightarrow{\vh}_{i}; \overleftarrow{\vh}_{i}]$.

\paragraph{Decoder}
The generation of each word $y_j$ is conditioned on all of the previously generated words $\vy_{<j}$ via the state of the RNN decoder $\vs_j$, and the source sentence via a \emph{dynamic} context vector $\vc_j$:

\vspace{-3mm}
{\small
\begin{eqnarray*}
y_j&\sim& \softmax(\vW_{y}\cdot \vr_j+\vb_r)\\
\vr_j&=& \tanh(\vs_j + \vW_{rc}\cdot \vc_j +\vW_{rj}\cdot \vE_T[y_{j-1}])\\
\vs_j&=& \tanh(\vW_{s}\cdot \vs_{j-1}+\vW_{sj}\cdot \vE_T[y_{j-1}] +\vW_{sc}\cdot \vc_j)\\
\end{eqnarray*}
}

\vspace{-8mm}
\hspace{-5mm} where $\vE_T[y_j]$ is embedding of the word $y_j$ from the embedding table $\vE_T$ of the target language, and $\vW$ matrices and $\vb_r$ vector are the parameters.
The dynamic context vector $\vc_j$  is computed via $\vc_j ~=~\sum_{i}\alpha_{ji}\vh_i$, where 
\begin{eqnarray*}
\valpha_j &=& \softmax(\va_j)\\
{a}_{ji}&=& \vv \cdot \tanh(\vW_{ae}\cdot \vh_i + \mathbf{W}_{at}\cdot \vs_{j-1})
\end{eqnarray*}
This is known as the \emph{attention} mechanism which  dynamically attends to relevant parts of the source  necessary for generating the next target word.

\subsection{Memory Networks (MemNets)}
Memory Networks \cite{Weston:15} are a class of neural models that use external memories to perform inference based on long-range dependencies. 
A memory is a collection of vectors $\vM=\{\vm_1,..,\vm_K\}$  constituting the memory cells, 
where each cell $\vm_k$ may potentially correspond to a discrete object $\vx_k$.
The memory is equipped with a \emph{read}  and optionally a \emph{write} operation.
Given a query vector $\vq$, the output vector {generated} by reading from the memory is $\sum_{i=1}^{|\vM|} p_i \vm_i$, 
where $p_i$ represents the relevance of the query to the $i$-th memory cell $\vp =\softmax(\vq^T\cdot \vM)$.
For the rest of the paper, we denote the read operation by $\MemNet(\vM,\vq)$.

\section{Document NMT as Structured Prediction}

\begin{figure}
\centering
  \includegraphics[width=0.7\linewidth]{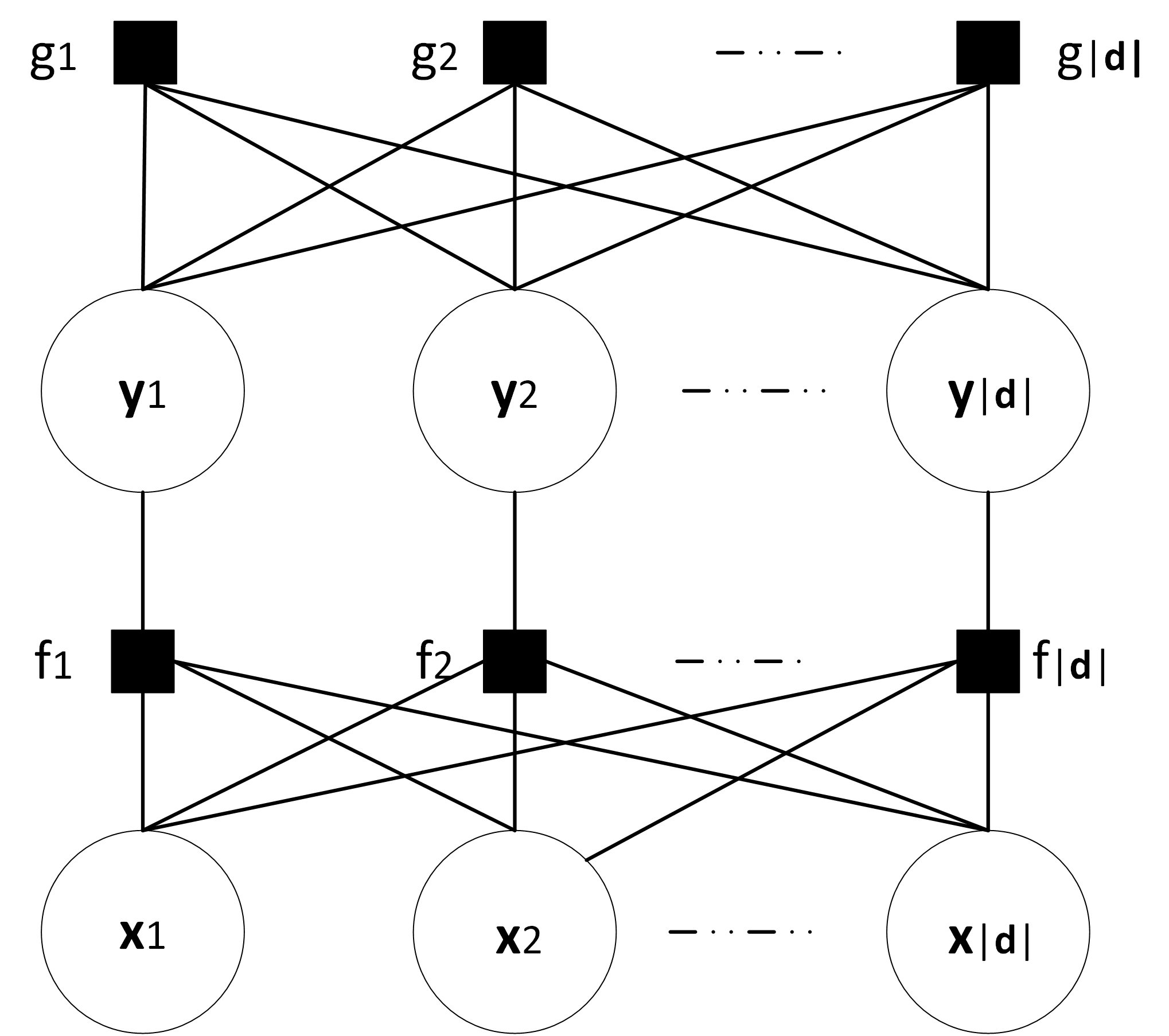}
\caption{Factor graph for document-level MT}
\label{fig:graph}
\end{figure}

We formulate document-wide machine translation as a \emph{structured} prediction problem. 
Given a set of  sentences $\{\vx_1, \ldots, \vx_{|\vd|}\}$ in a source document $\vd$, we are interested in generating the collection of their translations $\{\vy_1,\ldots,\vy_{|\vd|}\}$ taking into account \emph{interdependencies} among them imposed by the document.
 We achieve this by the factor graph  in Figure \ref{fig:graph} 
to model the  probability of  the target document given the source  document.  
 Our model has two types of factors:
\begin{itemize}
\item $f_{\vtheta}(\vy_t; \vx_t, \vx_{-t})$ to capture the interdependencies between the translation $\vy_t$, the corresponding source sentence $\vx_t$ and all the other sentences in the source document $\vx_{-t}$, and
\item $g_{\vtheta}(\vy_t; \vy_{-t})$ to capture the interdependencies between the translation $\vy_t$ and all  the other translations in the document $\vy_{-t}$.  
\end{itemize}
Hence, the probability of a document translation given the source document is
\begin{eqnarray}\label{eq:energy}
&& P(\vy_1,\ldots,\vy_{|\vd|}|\vx_1,\ldots,\vx_{|\vd|}) \propto \nonumber \\ 
&& \exp \Big(\sum_{t} f_{\vtheta}(\vy_t; \vx_t, \vx_{-t}) +  g_{\vtheta}(\vy_t; \vy_{-t})\Big). \nonumber
\end{eqnarray}
The factors $f_{\vtheta}$ and $g_{\vtheta}$ are realised by neural architectures whose parameters are collectively denoted by $\vtheta$.

\paragraph{Training} It is challenging to train the model parameters by maximising the (regularised) likelihood since computing the partition function is hard. 
This is due to the enormity of factors $g_{\vtheta}(\vy_t; \vy_{-t})$ over a large number of  translation variables $\vy_t$'s (i.e., the number of sentences in the document) as well as their unbounded domain (i.e., all sentences in the target language). 
Thus, we resort to maximising the \emph{pseudo-likelihood} \cite{Besag1975} for training the parameters:
\begin{eqnarray}
\arg\max_{\vtheta} \prod_{\vd \in \mathcal{D}}\prod_{t=1}^{|\vd|} P_{\vtheta}(\vy_t|\vx_t, \vy_{-t},\vx_{-t})  
\end{eqnarray}
where $\mathcal{D}$ is the set of bilingual training documents, and $|\vd|$ denotes the number of (bilingual) sentences in the document $\vd = \{(\vx_t,\vy_t)\}_{t=1}^{|\vd|}$. We directly model  the document-conditioned NMT model $P_{\vtheta}(\vy_t|\vx_t, \vy_{-t},\vx_{-t})$ using a neural architecture which subsumes both the $f_{\vtheta}$ and $g_{\vtheta}$ factors (covered in the next section). 

\paragraph{Decoding} To generate the best translation for a document according to our model, we need to solve the following optimisation problem:
$$\arg\max_{\vy_1,\ldots,\vy_{|\vd|}} \prod_{t=1}^{|\vd|} P_{\vtheta}(\vy_t|\vx_t, \vy_{-t},\vx_{-t}) $$
which is hard (due to similar reasons as mentioned earlier).
We hence resort to a block coordinate descent optimisation algorithm. More specifically, we initialise the  
translation of each sentence using the base neural MT model $P(\vy_t|\vx_t)$. 
We then repeatedly visit each sentence in the document, and update its translation using our document-context dependent NMT model $P(\vy_t|\vx_t, \vy_{-t},\vx_{-t}) $ while the translations of other sentences are kept fixed.

\section{Context Dependent NMT with MemNets}
\begin{figure}
 \centering
  \includegraphics[width=1.0\linewidth]{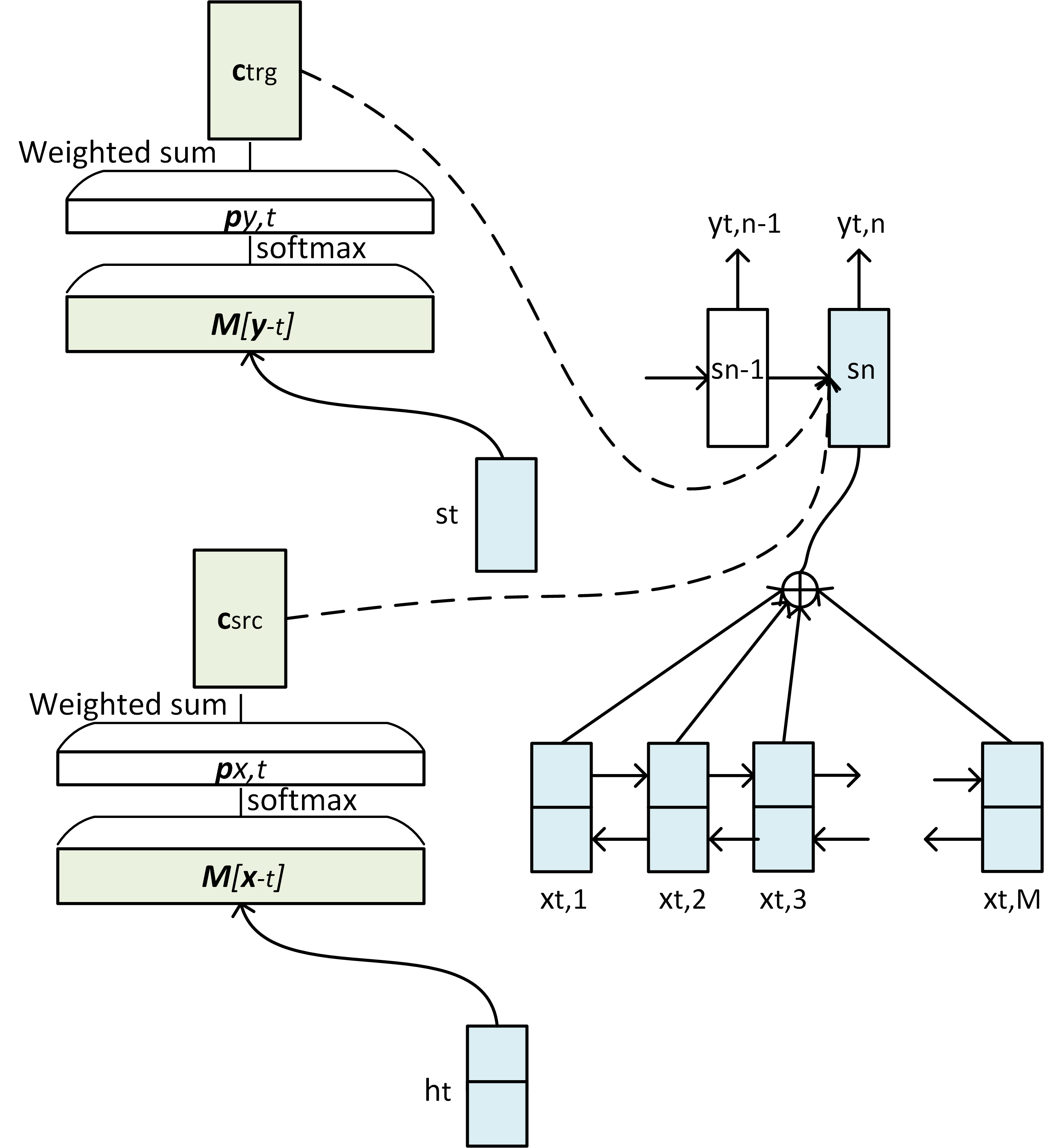}
 \caption{Our Memory-to-Context document-NMT model consisting of sentence-based NMT model with source and target external memories.}
\label{fig:docmt-memnn}
\end{figure}

We {augment} the sentence-level attentional NMT model by incorporating the document context (both source and target) using memory networks when generating the translation of a sentence, as shown in Figure~\ref{fig:docmt-memnn}. 

%
Our model generates the target translation word-by-word from left to right, similar to the vanilla attentional neural translation model.
However, it conditions the generation of a target word not only on the previously generated words and the current source sentence (as in the vanilla NMT model), 
but also on all the other source sentences of the document and their translations. That is, the generation process is as follows:
\begin{equation}\label{eq:objective}\small
P_{\vtheta}(\vy_{t} |\vx_t, \vy_{-t},\vx_{-t}) = \prod_{j=1}^{|\vy_t|} P_{\vtheta}(y_{t,j} |\vy_{t,<j}, \vx_t, \vy_{-t},\vx_{-t}) 
\end{equation}
where $y_{t,j}$ is the $j$-th word of the $t$-th  target sentence, $\vy_{t,<j}$ are the previously generated words, and $\vx_{-t}$ and $\vy_{-t}$ are {as introduced previously. }

Our model represents the source and target document contexts as external memories, and \emph{attends} to relevant 
parts of these external memories when generating the translation of a sentence.
Let $\vM[\vx_{-t}]$ and $\vM[\vy_{-t}]$ denote external memories representing the source and target document context, respectively.
These contain memory cells corresponding to all sentences in the document except the $t$-th sentence (described shortly). 
Let $\vh_t$ and $\vs_t$ be representations of the  $t$-th  source sentence and its current translation, from the encoder and decoder respectively. We make use of $\vh_t$ as the query to get the relevant \emph{context} from the source external memory:
$$\vc^{src}_t = \MemNet(\vM[\vx_{-t}], \vh_t)$$ 
Furthermore, for the $t$-th sentence, we get the relevant information from the target context:
$$\vc^{trg}_t = \MemNet(\vM[\vy_{-t}], \vs_t + \vW_{at} \cdot \vh_t)$$
where the query consists of the representation of the translation $\vs_t$ from the decoder endowed with that of the source sentence $\vh_t$ from the encoder to make the query robust to potential noises in the current translation and circumvent error propagation, and $\vW_{at}$ projects the source 
representation into the hidden state space.  

Now that we have representations of the relevant source and target document contexts, Eq.~\ref{eq:objective} can be re-written as:
\begin{equation}\small
P_{\vtheta}(\vy_{t} |\vx_t, \vy_{-t},\vx_{-t})  = \prod_{j=1}^{|\vy_t|} P_{\vtheta}(y_{t,j} |\vy_{t,<j}, \vx_t, \vc^{trg}_t,\vc^{src}_t) 
\end{equation}
%
%
More specifically, the memory contexts $\vc^{src}_t $ and $\vc^{trg}_t $ are incorporated into the NMT decoder as:
\begin{itemize}

\item {\bf Memory-to-Context} in which the memory contexts are incorporated when computing the next decoder hidden state:
\begin{eqnarray}
\mathbf{s}_{t,j} = \tanh({\vW}_{s}\cdot \mathbf{s}_{t,j-1}+{\vW}_{sj}\cdot \vE_T[y_{t,j}]+ \nonumber \\ 
{\vW}_{sc}\cdot {\vc}_{t,j}+ {\vW}_{sm}\cdot \vc_{t}^{src} +{\vW}_{st}\cdot \vc_{t}^{trg} ) \nonumber
\end{eqnarray}

\item  {\bf Memory-to-Output} in which the memory contexts are incorporated in the output layer:
\begin{eqnarray}
y_{t,j} &\sim& \softmax({\vW}_{y}\cdot \mathbf{r}_{t,j}+{\vW}_{ym}\cdot \vc_{t}^{src} + \nonumber \\
& & {\vW}_{yt}\cdot \vc_{t}^{trg} + \mathbf{b}_r) \nonumber
\end{eqnarray}

\end{itemize}
where ${\vW}_{sm}$, ${\vW}_{st}$, ${\vW}_{ym}$, and ${\vW}_{yt}$ are the new parameter matrices.
We use only the source, only the  target, or both external memories as the additional conditioning contexts. Furthermore, we use 
either the Memory-to-Context or Memory-to-Output architectures  for incorporating the document contexts. In the experiments, 
we will explore these different options to investigate the most effective combination. We now turn our attention to the construction of 
the external memories for the source and target sides of a document.


\paragraph{The Source  Memory}
We make use of a hierarchical 2-level RNN architecture to construct the external memory of the source document. 
More specifically, we pass each sentence of the document through a sentence-level bidirectional RNN to get the 
representation of the sentence (by concatenating the last hidden states of the forward and backward RNNs). 
We then pass the sentence representations through a document-level bidirectional RNN to propagate sentences' information  
across the document. We take the hidden states of the document-level bidirectional RNNs as the memory cells of the source external memory.

The source external memory is built once for each minibatch, and does not change throughout the document translation. 
To be able to fit the computational graph of the document NMT model within GPU memory limits, we pre-train 
the sentence-level bidirectional RNN  using the language modelling training objective. However, the document-level bidirectional RNN is trained  
together with other parameters of the document NMT model by back-propagating the document translation training objective. 

\paragraph{The Target  Memory}
 The memory cells of the target external memory represent the current translations of the document. 
 Recall  from the previous section that we use coordinate descent iteratively to update these translations. 
 Let $\{\vy_1,\ldots,\vy_{|\vd|}\}$ be the current translations, and let $\{\vs_{|\vy_1|},\ldots,\vs_{|\vy_{|\vd|}|}\}$ be the last states of the decoder 
 when these translations were generated. 
 We use these last decoder states as the cells of the external target memory. 
 We could make use of hierarchical sentence-document RNNs to transform the document translations into memory cells (similar to what we do for the source memory); however, it would have been computationally expensive and may have resulted in error propagation. 
We will show in the experiments that our efficient target memory construction is indeed effective.


\section{Experiments and Analysis}
\paragraph{Datasets.} We conducted experiments on three language pairs: French-English, German-English and Estonian-English. 
Table \ref{table:corpora} shows the statistics of the datasets used in our experiments. 
The French-English dataset is based on the TED Talks corpus\footnote{https://wit3.fbk.eu/} \cite{Cettolo:12} where each talk is {considered} a document. 
The Estonian-English data comes from the Europarl v7 corpus\footnote{http://www.statmt.org/europarl/} \cite{Koehn:05}. Following \newcite{Smith:13}, we split the speeches based on the \texttt{SPEAKER} tag and treat them as documents.
 {The French-English and Estonian-English corpora were randomly split into train/dev/test sets.} 
{For German-English, we use the News Commentary v9 corpus\footnote{http://statmt.org/wmt14/news-commentary-v9-by-document.tgz} for training, \texttt{news-dev2009} for  development, and  \texttt{news-test2011} and \texttt{news-test2016} as the test sets. 
The news-commentary corpus has document boundaries already provided.}

We pre-processed all corpora to remove very short documents and those with missing translations. Out-of-vocabulary and rare words (frequency less than 5) are replaced by the \unk  token, {following \newcite{Cohn:16}}.\footnote{{We do not split  words into subwords using BPE \cite{Sennrich:16} as that increases sentence lengths resulting in removing long documents due to GPU memory limitations, which would heavily reduce the amount of data that we have.}}

\setlength{\tabcolsep}{0.7pt}
\begin{table}
\centering
{\small
\begin{tabular}{l||c|c|c|c}
             & \# docs & \# sents & doc len & src/tgt vocab\\
\hline \hline
Fr-En & 10/1.2/1.5 & 123/15/19 & 123/128/124 & 25.1/21\\
Et-En & 150/10/18& 209/14/25& 14/14/14 & 48.6/24.9\\
De-En & 49/.9/1.1/1.6& 191/2/3/3& 39/23/27/19 & 45.1/34.7\\

\end{tabular}
}
\caption{Training/dev/test corpora statistics: number of documents ($\times$100) and sentences ($\times$1000), average document length (in sentences) and source/target vocabulary size ($\times$1000). For De-En, we report statistics of the two test sets \texttt{news-test2011} and \texttt{news-test2016}.}
\label{table:corpora}
\end{table}

\paragraph{Evaluation Measures}
We use BLEU \cite{Papineni:02} and METEOR \cite{Lavie:07} scores to measure the quality of the generated translations.
%
%
We use bootstrap resampling \cite{Clark:11} to measure statistical significance, \textit{p} $<$ 0.05,  comparing to  the baselines.

\paragraph{Implementation and Hyperparameters}
We implement our document-level neural machine translation model in C++ using the DyNet library \cite{dynet}, {on top of the basic sentence-level NMT implementation in \texttt{mantis} \cite{Cohn:16}}.    
{For the source memory, the sentence and document-level bidirectional RNNs use LSTM and GRU units, respectively. }
The translation model uses GRU units for the bidirectional RNN encoder and the 2-layer RNN decoder. 
GRUs are used instead of LSTMs to reduce the number of parameters in the {main model.}
The RNN hidden dimensions and word embedding sizes are set to 512 in the translation and memory components, and the alignment dimension is set to 256 in the translation model.

\setlength{\tabcolsep}{.8pt}
\begin{table*}
{\small
\centering
\begin{tabular}{l | c c c c | c c c c || c c c c | c c c c}
& \multicolumn{8}{c}{\textbf{Memory-to-Context}} & \multicolumn{8}{|c}{\textbf{Memory-to-Output}}\\
\hline                   
&\multicolumn{4}{c}{BLEU} &\multicolumn{4}{|c}{METEOR}  &\multicolumn{4}{|c}{BLEU}& \multicolumn{4}{|c}{METEOR}  \\  
& {Fr$\rightarrow$En} & \multicolumn{2}{c}{De$\rightarrow$En} & {Et$\rightarrow$En} & {Fr$\rightarrow$En} & \multicolumn{2}{c}{De$\rightarrow$En} & {Et$\rightarrow$En} & {Fr$\rightarrow$En} & \multicolumn{2}{c}{De$\rightarrow$En} & {Et$\rightarrow$En} & {Fr$\rightarrow$En} & \multicolumn{2}{c}{De$\rightarrow$En} & {Et$\rightarrow$En}\\
& & {\scriptsize NC-11} & {\scriptsize NC-16} & & & {\scriptsize NC-11} & {\scriptsize NC-16} & & & {\scriptsize NC-11} & {\scriptsize NC-16} & & & {\scriptsize NC-11} & {\scriptsize NC-16} & \\
\hline
\hline
\textit{S-NMT} & 20.85 & 5.24 & 9.18 & 20.42 & 23.27 & 10.90 & 14.35 & 24.65 & 20.85 & 5.24 & 9.18 & 20.42 & 23.27 & 10.90 & 14.35 & 24.65\\
\textit{\ \ +src} & 21.91$^\dagger$ & 6.26$^\dagger$ & 10.20$^\dagger$ & 22.10$^\dagger$ & 24.04$^\dagger$ & 11.52$^\dagger$ & 15.45$^\dagger$ & 25.92$^\dagger$ & \textbf{21.80}$^\dagger$ & 6.10$^\dagger$ & 9.98$^\dagger$ & 21.50$^\dagger$ & 23.99$^\dagger$ & 11.53$^\dagger$ & 15.29$^\dagger$ & 25.44$^\dagger$\\
\textit{\ \ +trg } & 21.74$^\dagger$ & 6.24$^\dagger$ & 9.97$^\dagger$ & 21.94$^\dagger$ & 23.98$^\dagger$ & 11.58$^\dagger$ & 15.32$^\dagger$ & 25.89$^\dagger$ & 21.76$^\dagger$ & \textbf{6.31}$^\dagger$ & 10.04$^\dagger$ & 21.82$^\dagger$ & 24.06$^\dagger$ & \textbf{12.10}$^\dagger$ & 15.75$^\dagger$ & 25.93$^\dagger$\\
\textit{\ \ +both} & \textbf{22.00}$^\dagger$ & \textbf{6.57}$^\dagger$ & \textbf{10.54}$^\dagger$ & \textbf{22.32}$^\dagger$ & \textbf{24.40}$^\dagger$ & \textbf{12.24}$^\dagger$ & \textbf{16.18}$^\dagger$& \textbf{26.34}$^\dagger$ & 21.77$^\dagger$ & 6.20$^\dagger$ & \textbf{10.23}$^\dagger$ & \textbf{22.20}$^\dagger$ & \textbf{24.27}$^\dagger$ & 11.84$^\dagger$ & \textbf{15.82}$^\dagger$ & \textbf{26.10}$^\dagger$\\
\hline                          
\end{tabular}
}
\caption{BLEU and METEOR scores for the sentence-level baseline (S-NMT) vs. variants of our Document NMT model. \textbf{bold}: Best performance, $\dagger$: Statistically significantly better than the baseline.}
\label{table:bleu}
\end{table*}

\paragraph{Training} 
We use a stage-wise method to train the variants of our document context NMT model. 
Firstly, we pre-train the Memory-to-Context/Memory-to-Output models, setting their \emph{readings} from the source and target memories to the zero vector. 
This effectively learns parameters associated with the underlying sentence-based NMT model, which is then used 
as initialisation when training \emph{all} parameters in the second stage (including the ones from the first stage).
For the first stage, we make use of stochastic gradient descent (SGD)\footnote{{In our initial experiments, we found SGD to be more effective than Adam/Adagrad;  an  observation also made by  \newcite{Bahar:17}.}} with initial learning rate of 0.1 and a decay factor of 0.5 after the fourth epoch for a total of ten epochs. {The convergence occurs in 6-8 epochs.}
For the second stage, we use SGD with an initial learning rate of 0.08 and  a decay factor of 0.9 after the first epoch for 
a total of 15 epochs\footnote{\sameen{For the document NMT model training, we did some preliminary experiments using different learning rates and used the scheme which converged to the best perplexity in the least number of epochs while for sentence-level training we follow \newcite{Cohn:16}.}}. The best model is picked based on the dev-set perplexity.
To avoid overfitting, we employ dropout with the rate 0.2 for the single memory model. For the dual memory model, we set dropout for Document RNN to 0.2 and for the encoder and decoder to 0.5. Mini-batching is used in both stages to speed up training. \sameen{For the largest dataset, the document NMT model takes about 4.5 hours per epoch to train on a single P100 GPU, while the sentence-level model takes about 3 hours per epoch for the same settings.}

When training the document NMT model in the second stage, we need the target memory.
One option would be to use the ground truth  translations for building the memory. 
However, this may result in inferior training, since at the test time, the decoder {iteratively} updates the translation of sentences based on the  noisy translations of other sentences (accessed via the target memory). 
Hence, while training the document NMT model, we construct the target memory from the translations \emph{generated} by the pre-trained sentence-level model\footnote{\sameen{We report results for two-pass decoding, i.e., we only update the translations once using the initial translations generated from the base model. We tried multiple passes of decoding at test-time but it was not helpful.}}.
 This effectively exposes the model to its potential test-time mistakes during the training time, resulting in more robust learned parameters.
 %
 

\subsection{Main Results}
We have three variants of our model, using: (i) only the source memory (\srcmem), (ii) only the target memory (\trgmem), or (iii) both the source and target memories (\bothmem). We compare these variants against the standard sentence-level NMT model (\snmt). We also compare the source memory  variants of our model to the local context-NMT models\footnote{We implemented and trained the baseline local context models using the same hyperparameters and training procedure that we used for training our  memory models.} of \newcite{Jean:17} and \newcite{Wang:17}, which use a few previous source sentences as context, added to the decoder hidden state (similar to our Memory-to-Context model).



\paragraph{Memory-to-Context} 
We consistently observe +1.15/+1.13  BLEU/METEOR score improvements across the three language pairs upon comparing our best model to \snmt \ (see Table \ref{table:bleu}). 
Overall, our document NMT model  with both memories has been the most effective variant for all of the three language pairs. 
%

We further experiment to train the target memory variants using \emph{gold} translations instead of the generated ones for German-English. 
This led to $-0.16$ and $-0.25$ decrease\footnote{Latter is statistically significant decrease w.r.t. the both memory model trained on generated target translations.} in the BLEU scores for the target-only and both-memory variants, which confirms the intuition of constructing the target memory by exposing the model to its noises during training time. 
%


\setlength{\tabcolsep}{1pt}
\begin{table}\small
\centering
\begin{tabular}{l | c c c | c c c }
& \multicolumn{3}{c}{\textbf{Memory-to-Context}} & \multicolumn{3}{|c}{\textbf{Memory-to-Output}}\\
{Lang. Pair} & {\scriptsize Fr$\rightarrow$En} & {\scriptsize De$\rightarrow$En} & {\scriptsize Et$\rightarrow$En} & {\scriptsize Fr$\rightarrow$En} & {\scriptsize De$\rightarrow$En} & {\scriptsize Et$\rightarrow$En}\\
\hline \hline
\textit{S-NMT} & 42.5 & 66.8 & 58.4 & 42.5 & 66.8 & 58.5\\
\textit{\  +src mem} & 48.8 & 73.1 & 64.8 & 68.7 & 107.1 & 88.7\\
\textit{\  +trg mem} & 43.8 & 68.1 & 59.8 & 53.8 & 85.1 & 71.8\\
\textit{\  +both mems} & 50.1 & 74.4 & 66.1 & 80 & 125.4 & 102\\
\end{tabular}
\caption{Number of model parameters (millions).}
\label{table:param}
\end{table}

\pgfplotstableread[row sep=\\,col sep=&]{
    size & base & src & trg & both\\
    Smaller Corpus & 10.90 & 11.53 & 12.10 & 11.84\\
    Larger Corpus & 12.12 & 12.48 & 13.21 & 12.99\\
    }\mydata

\begin{figure}[h]
\centering
\subcaptionbox{Memory-to-Context model}
{
\begin{tikzpicture}
    \begin{axis}[
            ybar,
	enlarge x limits=0.5,
            x=3cm,
            bar width=0.5cm,
            height=.7\linewidth,
            legend style={at={(0.5,1)},
                anchor=north,legend columns=-1},
            symbolic x coords={Smaller Corpus, Larger Corpus},
            xtick={Smaller Corpus, Larger Corpus},
	every node near coord/.append style={font=\scriptsize},
            nodes near coords,
            nodes near coords align={vertical},
            ymin=10,ymax=15.0,
            ylabel={METEOR},
        ]
        \addplot coordinates{(Smaller Corpus, 10.90) (Larger Corpus, 12.12)};
        \addplot coordinates{(Smaller Corpus, 11.52) (Larger Corpus, 12.94)};
        \addplot coordinates{(Smaller Corpus, 11.58) (Larger Corpus, 12.55)};
        \addplot coordinates{(Smaller Corpus, 12.24) (Larger Corpus, 13.56)};
        \legend{\scriptsize{S-NMT}, \scriptsize{S-NMT+src},\scriptsize{S-NMT+trg},\scriptsize{S-NMT+both}}
    \end{axis}
\end{tikzpicture}
}
\vspace{0.35cm}

\subcaptionbox{Memory-to-Output model}
{
\begin{tikzpicture}
    \begin{axis}[
            ybar,
	enlarge x limits=0.5,
            x=3cm,
            bar width=0.5cm,
            height=.7\linewidth,
            legend style={at={(0.5,1)},
                anchor=north,legend columns=-1},
            symbolic x coords={Smaller Corpus, Larger Corpus},
            xtick={Smaller Corpus, Larger Corpus},
	every node near coord/.append style={font=\scriptsize},
            nodes near coords,
            nodes near coords align={vertical},
            ymin=10,ymax=14.5,
            ylabel={METEOR},
        ]
        \addplot coordinates{(Smaller Corpus, 10.90) (Larger Corpus, 12.12)};
        \addplot coordinates{(Smaller Corpus, 11.53) (Larger Corpus, 12.48)};
        \addplot coordinates{(Smaller Corpus, 12.10) (Larger Corpus, 13.21)};
        \addplot coordinates{(Smaller Corpus, 11.84) (Larger Corpus, 12.99)};
        \legend{\scriptsize{S-NMT}, \scriptsize{S-NMT+src},\scriptsize{S-NMT+trg},\scriptsize{S-NMT+both}}
    \end{axis}
\end{tikzpicture}
}

\caption{METEOR scores on De$\rightarrow$En (NC-11) while training S-NMT with smaller vs. larger corpus.}\label{fig:compare}
\end{figure}
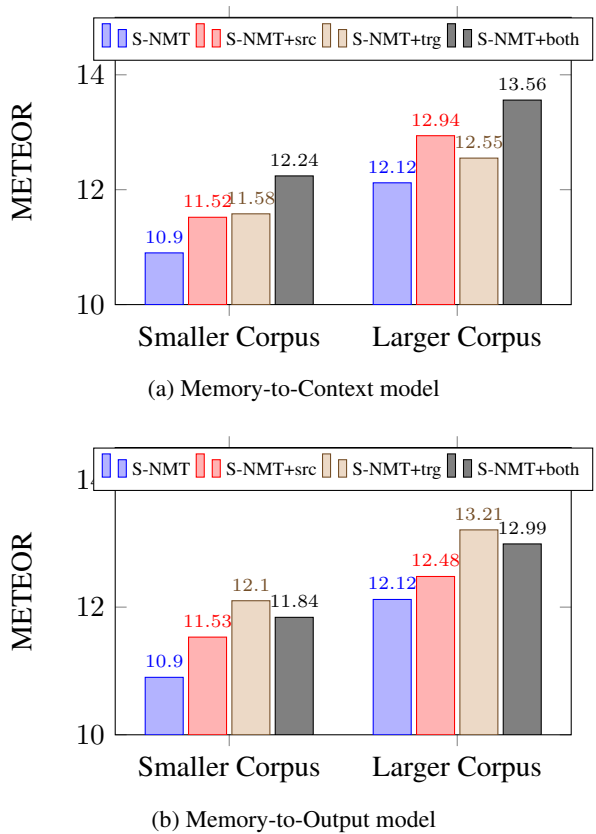

\pgfplotstableread[row sep=\\,col sep=&]{
    size & base & src & trg & both\\
    Smaller Corpus & 10.90 & 11.52 & 11.58 & 12.24\\
    Larger Corpus & 12.12 & 12.94 & 12.55 & 13.56\\
    }\myctx

\setlength{\tabcolsep}{0.3pt}

\begin{table*}[thb]
\begin{minipage}{0.58\linewidth}
{\small
\begin{tabular}{l | c c c c | c c c c}
&\multicolumn{4}{c}{BLEU} &\multicolumn{4}{|c}{METEOR} \\  
& Fr$\rightarrow$En & \multicolumn{2}{c}{De$\rightarrow$En} & Et$\rightarrow$En & Fr$\rightarrow$En & \multicolumn{2}{c}{De$\rightarrow$En} & Et$\rightarrow$En\\
& & {\scriptsize NC-11} & {\scriptsize NC-16} & & & {\scriptsize NC-11} & {\scriptsize NC-16} & \\
\hline
\hline
{\scriptsize \newcite{Jean:17}} & 21.95 & 6.04 & 10.26 & 21.67 & 24.10 & 11.61 & 15.56 & 25.77\\
{\scriptsize \newcite{Wang:17}} & 21.87 & 5.49 & 10.14 & 22.06 & 24.13 & 11.05 & 15.20 & 26.00\\
\hline
\textit{S-NMT} & 20.85 & 5.24 & 9.18 & 20.42 & 23.27 & 10.90 & 14.35 & 24.65\\
\ \ \textit{+src mem} & 21.91$^\dagger$ & 6.26$^\clubsuit$ & 10.20 & 22.10$^\spadesuit$ & 24.04$^\dagger$ & 11.52$^\clubsuit$ & 15.45$^\clubsuit$ & 25.92$^\spadesuit$\\
\ \ \textit{+both mems} & \textbf{22.00}$^\dagger$ & \textbf{6.57}$^{\diamondsuit}$ & \textbf{10.54}$^\clubsuit$ & \textbf{22.32}$^{\diamondsuit}$ & \textbf{24.40}$^\diamondsuit$ & \textbf{12.24}$^{\diamondsuit}$ & \textbf{16.18}$^{\diamondsuit}$ & \textbf{26.34}$^{\diamondsuit}$\\
\hline
\end{tabular}
}
\caption{Our Memory-to-Context Source Memory NMT variants vs. S-NMT and Source context NMT baselines. \textbf{bold}: Best performance, $\dagger$, $\spadesuit$, $\clubsuit$, $\diamondsuit$: Statistically significantly better than only S-NMT, S-NMT \& \newcite{Jean:17}, S-NMT \& \newcite{Wang:17}, all baselines, respectively.}
\label{table:bleu-aux}
\end{minipage}
\hfill
\begin{minipage}{.37\linewidth}
{\small
\begin{tabular}{l | c c c c}
&\multicolumn{4}{c}{BLEU-1}\\  
& {Fr$\rightarrow$En} & \multicolumn{2}{c}{De$\rightarrow$En} & {Et$\rightarrow$En} \\
& & {\scriptsize NC-11} & {\scriptsize NC-16} & \\
\hline
\hline
{\scriptsize \newcite{Jean:17}} & 52.8 & 30.6 & 39.2 & 51.9 \\
{\scriptsize \newcite{Wang:17}} & 52.6 & 28.2 & 38.3 & 52.3 \\
\hline
\textit{S-NMT} & 51.4 & 28.7 & 36.9 & 50.4\\
\ \ \textit{+src mem} & 53.0 & 30.5 & 39.1 & 52.6\\
\ \ \textit{+both mems} & \textbf{53.5} & \textbf{33.1} & \textbf{41.3} & \textbf{53.2}\\
\hline
\end{tabular}
}
\caption{Unigram BLEU for our Memory-to-Context Document NMT models vs. S-NMT and Source context NMT baselines. \textbf{bold}: Best performance.}
\label{table:unibleu-aux}
	\end{minipage}
\end{table*}

\paragraph{Memory-to-Output} From Table \ref{table:bleu}, 
we consistently see +.95/+1.00 BLEU/METEOR improvements between the best variants of our model and the sentence-level baseline across the three language pairs. 
For French$\rightarrow$English, all variants of document NMT model show comparable performance {when using BLEU; however, when evaluated using METEOR, the dual memory model is the best}. 
For German$\rightarrow$English, the target memory variants give comparable results, whereas for Estonian$\rightarrow$English, the dual memory variant proves to be the best. 
%
%
Overall, the Memory-to-Context model variants perform better than their Memory-to-Output counterparts.
 We attribute this to the large number of parameters in the latter architecture (Table~\ref{table:param}) and limited amount of data. 

 We further experiment with more data for training the sentence-based NMT  to investigate the extent to which document context  is  useful in this setting.  
 We randomly choose an additional 300K German-English sentence pairs from WMT'14 data to train the  base NMT  model in stage 1. 
In stage 2, we use the same document corpus as before to train the  document-level models.
As seen from Figure~\ref{fig:compare}, the document MT variants still benefit from the document context even when the base model is trained on a larger bilingual corpus. 
{\sameen{For the Memory-to-Context model, we see massive improvements of $+0.72$ and $+1.44$ METEOR scores for the source memory and dual memory model respectively, when compared to the baseline.} On the other hand, for the Memory-to-Output model, the target memory model's METEOR score increases significantly by $+1.09$  compared to the baseline, slightly differing from the corresponding model using the smaller corpus ($+1.2$).}

\paragraph{Local Source Context Models} Table~\ref{table:bleu-aux} shows comparison of our Memory-to-Context model variants to local source context-NMT models \cite{Jean:17,Wang:17}. For French$\rightarrow$English, our source memory model is comparable to both baselines. For German$\rightarrow$English, our \textit{S-NMT+src mem} model is comparable to \newcite{Jean:17} but outperforms \newcite{Wang:17} {for one test set according to  BLEU, and for both test sets according to  METEOR}. For Estonian$\rightarrow$English, our model outperforms \newcite{Jean:17}. 
Our global source context model has only surface-level sentence information, and is oblivious to the individual words in the context since we do an offline training to get the sentence representations (as previously mentioned).  However, the other two context baselines have access to that information, yet  our model's performance is either better or quite close to those models. 
We also look into the unigram BLEU scores to see how much our global source memory variants lead to improvement at the word-level. 
From Table~\ref{table:unibleu-aux}, it can be seen that our model's performance is better than the baselines for majority of the cases. 
%
The \textit{S-NMT+both mems} model gives the best results for all three language pairs, showing that leveraging both source and target document context is indeed beneficial for improving MT performance. 

\subsection{Analysis}

\setlength{\tabcolsep}{1pt}

\begin{table}[b]
\centering
{\small
\begin{tabular}{l | c c c | c c c}
&\multicolumn{3}{c}{BLEU} &\multicolumn{3}{|c}{METEOR} \\  
{Lang. Pair}& {\scriptsize Fr$\rightarrow$En} & {\scriptsize De$\rightarrow$En} & {\scriptsize Et$\rightarrow$En} & {\scriptsize Fr$\rightarrow$En} & {\scriptsize De$\rightarrow$En} & {\scriptsize Et$\rightarrow$En}\\
\hline
\textit{S-NMT} & 20.85 & 5.24 & 20.42 & 23.27 & 10.90 & 24.65\\
\ \ \textit{+prev trg} & \textbf{21.75} & 5.93 & \textbf{22.08} & \textbf{24.03} & 11.40 & \textbf{25.94}\\
\ \ \textit{+trg mem} & 21.74 & \textbf{6.24} & 21.94 & 23.98 & \textbf{11.58} & 25.89\\
\hline
\end{tabular}
}
\caption{Analysis of target context model.}
\label{table:bleu-trg}
\end{table}


\paragraph{Using Global/Local Target Context}
We first investigate whether using a local target context would have been equally sufficient in comparison to our global target memory model for the three datasets. We condition the decoder on the previous target sentence representation (obtained from the last hidden state of the decoder) by 
adding it as an additional input to all decoder states (\textit{PrevTrg}) similar to our Memory-to-Context model. 
From Table~\ref{table:bleu-trg}, we observe that for French$\rightarrow$English and Estonian$\rightarrow$English, using all sentences in the target context or just the previous target sentence gives comparable results. 
{We may attribute this to these specific datasets, that is documents from TED talks or European Parliament Proceedings may depend more on the local than on the global context.}
However, for German$\rightarrow$English (NC-11), the target memory model performs the best showing that for documents with richer context (e.g. news articles) we do need the global target document context to improve MT performance.

\paragraph{Output Analysis}

To better understand the dual memory model, we look at the first sentence example in Table~\ref{table:example}. 
It can be seen that the source sentence has the noun ``Qimonda" but the sentence-level NMT model fails to attend to it when generating the translation. On the other hand, the single memory models are better in delivering some, if not all, of the underlying information in the source sentence but the dual memory model's translation quality surpasses them. \sameen{This is because the word ``Qimonda'' was being repeated in this specific document, providing a strong contextual signal to our global document context model while} the local context model by \newcite{Wang:17} is still unable to correctly translate the noun even when it has access to the word-level information of  previous sentences.

We resort to manual evaluation as there is no standard metric which evaluates document-level discourse information like consistency or pronominal anaphora. By manual inspection, we observe that our models can identify nouns in the source sentence to resolve coreferent pronouns, as shown in the second example of Table~\ref{table:example}. Here the topic of the sentence is ``\textit{the country under the dictatorship of Lukashenko}'' and our target and dual memory models are able to generate the appropriate pronoun/determiner as well as accurately translate the word `\textit{diktatuur}', hence producing much better translation as compared to both baselines. Apart from these improvements, our models are better in improving the readability of sentences by generating more context appropriate grammatical structures such as verbs and adverbs.

Furthermore, to validate that our model improves the consistency of translations, we look at five documents (roughly 70 sentences) from the test set of Estonian-English, each of which had a word being repeated in the gold translation. Our model is able to resolve the consistency in 22 out of 32 cases as compared to the sentence-based model which only accurately translates 16 of those. Following \newcite{Wang:17}, we also  investigate the extent to which our model can correct errors made by the baseline system. We randomly choose five documents from the test set. Out of the 20 words/phrases which were incorrectly translated by the sentence-based model, our model corrects 85\% of them while also generating 10\% new errors. 



\setlength{\tabcolsep}{2.5pt}
\begin{table}[th]
\centering
{\scriptsize
\begin{tabular}{|l|l|}
\hline
\textit{Source} & {\color{blue}qimonda} t\"aidab lissaboni strateegia eesm\"arke. \\
\textit{Target} & {\color{blue}qimonda} meets the objectives of the lisbon strategy.\\
\hline
\textit{S-NMT} & {\color{red}\unk} is the objectives of the lisbon strategy. \\
\textit{ +Src Mem} & {\color{red}the millennium development goals} are fulfilling the\\
 & \ \  \ millennium goals of the lisbon strategy. \\
\textit{\ +Trg Mem} & in writing. - (ro) {\color{red}the lisbon strategy} is fulfilling the\\
& \ \ \  objectives of the lisbon strategy. \\
\textit{\ +Both Mems} & {\color{blue}qimonda} fulfils the aims of the lisbon strategy.\\
\hline
{\scriptsize \newcite{Wang:17}} & {\color{red}\unk} fulfils the objectives of the lisbon strategy. \\
\hline \hline                  
\textit{Source} & ...
et riigis kehtib {\color{blue} endiselt luka\v senka diktatuur},\\
 & \ \ \ mis rikub inim- ning etnilise v\"ahemuse \~oigusi. \\
\textit{Target} & ...
this country is still {\color{blue}under the dictatorship of}\\
 & \ \ \ {\color{blue}lukashenko,} breaching human rights and the rights\\
 & \ \ \ of ethnic minorities. \\
\hline
\textit{S-NMT} & ...
the country still {\color{red} remains in a position of lukashenko}\\
 & \ \ \ {\color{red} to} violate human rights and ethnic minorities.\\
\textit{\ +Src Mem} & ...
the country still {\color{red}applies to the brutal dictatorship of}\\
 & \ \ \  human and ethnic minority rights.\\
\textit{\ +Trg Mem} & ...
the country still {\color{blue}keeps the \unk dictatorship that}\\
 & \ \ \ violates human rights and ethnic rights. \\
\textit{\ +Both Mems} & ...
the country still {\color{blue}persists in lukashenko's dictatorship}\\
 & \ \ \ {\color{blue} that} violate human rights and ethnic minority rights. \\
\hline
{\scriptsize \newcite{Wang:17}} & ... there is still a {\color{red}regime in the country that} is \\
& \ \ \ violating the rights of human and ethnic minority\\
& \ \ \  in the country.\\
\hline
\end{tabular}
}
\caption{Example Et$\rightarrow$En sentence translations (Memory-to-Context) from two test documents.}
\label{table:example}
\end{table}

\section{Related Work}
\paragraph{Document-level Statistical MT} 
 There have been a few SMT-based attempts to document MT, but they are either restrictive or do not lead to significant improvements. 
\newcite{Hardmeier:10} identify links among words in the source document  using a word-dependency model to improve translation of anaphoric pronouns. 
\newcite{Gong:11} make use of  a cache-based system to save relevant information from the previously generated  translations and use that to enhance document-level translation. 
\newcite{Garcia:14} propose a two-pass approach to improve the translations  already obtained by a sentence-level model. 

Docent is  an SMT-based document-level decoder \cite{Hardmeier:12,hardmeier-EtAl:2013:SystemDemo}, which tries to modify the initial translation generated by the Moses decoder \cite{Koehn:07} through stochastic local search and hill-climbing. 
\newcite{Martinez:15} make use of neural-based  continuous word representations to incorporate distributional semantics into Docent. In another work, \newcite{Garcia:17} incorporate new word embedding features into Docent to improve the lexical consistency of translations. The proposed methods fail to yield improvements upon automatic evaluation.

\paragraph{Larger Context Neural MT} 

\newcite{Jean:17} extend the vanilla attention-based neural MT model  \cite{Bahdanau:15} by conditioning the decoder on the previous sentence via attention over its words. Extending their model to consider the global source document context would be challenging due to the large size of  computation graph over all the words in the source document. 
\newcite{Wang:17} employ a 2-level hierarichal  RNN to summarise three previous source sentences, which is then used as an additional input to the decoder hidden state. 
\newcite{Bawden:17} use multi-encoder NMT models to exploit context from the previous source and target sentence. They highlight the importance of target-side context but report deteriorated BLEU scores when using it. All these works consider a very local source/target context and completely ignore the global source and target document contexts. 



\section{Conclusion}
We have proposed a document-level neural MT model that captures global source and target document context.
Our model augments the vanilla sentence-based NMT model with external memories to incorporate 
documental interdependencies on both source and target sides. 
%
We show statistically significant improvements of the translation quality on  three language pairs. 
For future work, we intend to investigate models which incorporate specific discourse-level phenomena.

\section*{Acknowledgments}
The authors are grateful to Andr{\'e} Martins and the anonymous reviewers for their helpful comments and corrections. This work was supported by the Multi-modal Australian ScienceS Imaging and Visualisation Environment (MASSIVE) (\url{www.massive.org.au}), and partially supported by a Google Faculty Award to GH and  the Australian Research Council through DP160102686. 

\bibliography{acl2018}

\begin{thebibliography}{29}
\expandafter\ifx\csname natexlab\endcsname\relax\def\natexlab#1{#1}\fi

\bibitem[{Bahar et~al.(2017)Bahar, Alkhouli, Peter, Brix, and Ney}]{Bahar:17}
Parnia Bahar, Tamer Alkhouli, Jan-Thorsten Peter, Christopher Jan-Steffen Brix,
  and Hermann Ney. 2017.
\newblock Empirical investigation of optimization algorithms in neural machine
  translation.
\newblock In \emph{Conference of the European Association for Machine
  Translation}, pages 13--26, Prague, Czech Republic.

\bibitem[{Bahdanau et~al.(2015)Bahdanau, Cho, and Bengio}]{Bahdanau:15}
Dzmitry Bahdanau, Kyunghyun Cho, and Yoshua Bengio. 2015.
\newblock Neural machine translation by jointly learning to align and
  translate.
\newblock In \emph{Proceedings of the International Conference on Learning
  Representations}.

\bibitem[{Bawden et~al.(2017)Bawden, Sennrich, Birch, and Haddow}]{Bawden:17}
Rachel Bawden, Rico Sennrich, Alexandra Birch, and Barry Haddow. 2017.
\newblock Evaluating discourse phenomena in neural machine translation.
\newblock In \emph{arXiv:1711.00513}.

\bibitem[{Besag(1975)}]{Besag1975}
Julian Besag. 1975.
\newblock Statistical analysis of non-lattice data.
\newblock \emph{Journal of the Royal Statistical Society. Series D (The
  Statistician)}, 24(3):179--195.

\bibitem[{Cettolo et~al.(2012)Cettolo, Girardi, and Federico}]{Cettolo:12}
Mauro Cettolo, Christian Girardi, and Marcello Federico. 2012.
\newblock {WIT}$^3$: Web inventory of transcribed and translated talks.
\newblock In \emph{Proceedings of the 16$^{th}$ Conference of the European
  Association for Machine Translation}, pages 261--268.

\bibitem[{Cho et~al.(2014)Cho, {van Merrienboer}, Bahdanau, and
  Bengio}]{cho_gru2014}
Kyunghyun Cho, B~{van Merrienboer}, Dzmitry Bahdanau, and Yoshua Bengio. 2014.
\newblock On the properties of neural machine translation: Encoder-decoder
  approaches.
\newblock In \emph{Eighth Workshop on Syntax, Semantics and Structure in
  Statistical Translation (SSST-8)}.

\bibitem[{Clark et~al.(2011)Clark, Dyer, Lavie, and Smith}]{Clark:11}
Jonathan~H. Clark, Chris Dyer, Alon Lavie, and Noah~A. Smith. 2011.
\newblock \href {http://www.aclweb.org/anthology/P11-2031} {Better hypothesis
  testing for statistical machine translation: Controlling for optimizer
  instability}.
\newblock In \emph{Proceedings of the 49th Annual Meeting of the Association
  for Computational Linguistics: Human Language Technologies (Short Papers)},
  pages 176--181. Association for Computational Linguistics.

\bibitem[{Cohn et~al.(2016)Cohn, Hoang, Vymolova, Yao, Dyer, and
  Haffari}]{Cohn:16}
Trevor Cohn, Cong Duy~Vu Hoang, Ekaterina Vymolova, Kaisheng Yao, Chris Dyer,
  and Gholamreza Haffari. 2016.
\newblock \href {http://www.aclweb.org/anthology/N16-1102} {Incorporating
  structural alignment biases into an attentional neural translation model}.
\newblock In \emph{Proceedings of the North American Chapter of the Association
  for Computational Linguistics: Human Language Technologies}, pages 876--885.
  Association for Computational Linguistics.

\bibitem[{Garcia et~al.(2017)Garcia, Creus, Espa{\~n}a-Bonet, and
  M{\`a}rquez}]{Garcia:17}
Eva~Mart{\'i}nez Garcia, Carles Creus, Cristina Espa{\~n}a-Bonet, and Llu{\'i}s
  M{\`a}rquez. 2017.
\newblock Using word embeddings to enforce document-level lexical consistency
  in machine translation.
\newblock \emph{The Prague Bulletin of Mathematical Linguistics}, 108:85--96.

\bibitem[{Garcia et~al.(2014)Garcia, Espa{\~n}a-Bonet, and
  M{\`a}rquez}]{Garcia:14}
Eva~Mart{\'i}nez Garcia, Cristina Espa{\~n}a-Bonet, and Llu{\'i}s M{\`a}rquez.
  2014.
\newblock Document-level machine translation as a re-translation process.
\newblock \emph{Procesamiento del Lenguaje Natural}, 53:103--110.

\bibitem[{Garcia et~al.(2015)Garcia, Espa{\~n}a-Bonet, and
  M{\`a}rquez}]{Martinez:15}
Eva~Mart{\'i}nez Garcia, Cristina Espa{\~n}a-Bonet, and Llu{\'i}s M{\`a}rquez.
  2015.
\newblock Document-level machine translation with word vector models.
\newblock In \emph{Proceedings of the18th Conference of the European
  Association for Machine Translation}, pages 59--66.

\bibitem[{Gong et~al.(2011)Gong, Zhang, and Zhou}]{Gong:11}
Zhengxian Gong, Min Zhang, and Guodong Zhou. 2011.
\newblock \href {http://dl.acm.org/citation.cfm?id=2145432.2145532}
  {Cache-based document-level statistical machine translation}.
\newblock In \emph{Proceedings of the Conference on Empirical Methods in
  Natural Language Processing}, pages 909--919. Association for Computational
  Linguistics.

\bibitem[{Hardmeier and Federico(2010)}]{Hardmeier:10}
Christian Hardmeier and Marcello Federico. 2010.
\newblock Modelling pronominal anaphora in statistical machine translation.
\newblock In \emph{International Workshop on Spoken Language Translation},
  pages 283--289.

\bibitem[{Hardmeier et~al.(2012)Hardmeier, Nivre, and Tiedemann}]{Hardmeier:12}
Christian Hardmeier, Joakim Nivre, and J\"{o}rg Tiedemann. 2012.
\newblock \href {http://www.aclweb.org/anthology/D12-1108} {Document-wide
  decoding for phrase-based statistical machine translation}.
\newblock In \emph{Proceedings of the 2012 Joint Conference on Empirical
  Methods in Natural Language Processing and Computational Natural Language
  Learning}, pages 1179--1190. Association for Computational Linguistics.

\bibitem[{Hardmeier et~al.(2013)Hardmeier, Stymne, Tiedemann, and
  Nivre}]{hardmeier-EtAl:2013:SystemDemo}
Christian Hardmeier, Sara Stymne, J\"{o}rg Tiedemann, and Joakim Nivre. 2013.
\newblock \href {http://www.aclweb.org/anthology/P13-4033} {Docent: A
  document-level decoder for phrase-based statistical machine translation}.
\newblock In \emph{Proceedings of the 51st Annual Meeting of the Association
  for Computational Linguistics: System Demonstrations}, pages 193--198.
  Association for Computational Linguistics.

\bibitem[{Hochreiter and Schmidhuber(1997)}]{Hochreiter:97}
Sepp Hochreiter and J\"{u}rgen Schmidhuber. 1997.
\newblock Long short-term memory.
\newblock \emph{Neural Comput.}, 9(8):1735--1780.

\bibitem[{Jean et~al.(2017)Jean, Lauly, Firat, and Cho}]{Jean:17}
Sebastien Jean, Stanislas Lauly, Orhan Firat, and Kyunghyun Cho. 2017.
\newblock Does neural machine translation benefit from larger context?
\newblock In \emph{arXiv:1704.05135}.

\bibitem[{Koehn(2005)}]{Koehn:05}
Philipp Koehn. 2005.
\newblock Europarl: A parallel corpus for statistical machine translation.
\newblock In \emph{Conference Proceedings: the 10th Machine Translation
  Summit}, pages 79--86. AAMT.

\bibitem[{Koehn et~al.(2007)Koehn, Hoang, Birch, Callison-Burch, Federico,
  Bertoldi, Cowan, Shen, Moran, Zens, Dyer, Bojar, Constantin, and
  Herbst}]{Koehn:07}
Philipp Koehn, Hieu Hoang, Alexandra Birch, Chris Callison-Burch, Marcello
  Federico, Nicola Bertoldi, Brooke Cowan, Wade Shen, Christine Moran, Richard
  Zens, Chris Dyer, Ond\v{r}ej Bojar, Alexandra Constantin, and Evan Herbst.
  2007.
\newblock \href {http://www.aclweb.org/anthology/P07-2045} {Moses: Open source
  toolkit for statistical machine translation}.
\newblock In \emph{Proceedings of the 45th Annual Meeting of the ACL on
  Interactive Poster and Demonstration Sessions}, pages 177--180. Association
  for Computational Linguistics.

\bibitem[{Lavie and Agarwal(2007)}]{Lavie:07}
Alon Lavie and Abhaya Agarwal. 2007.
\newblock \href {http://dl.acm.org/citation.cfm?id=1626355.1626389} {Meteor: An
  automatic metric for mt evaluation with high levels of correlation with human
  judgments}.
\newblock In \emph{Proceedings of the Second Workshop on Statistical Machine
  Translation}, StatMT '07, pages 228--231, Stroudsburg, PA, USA. Association
  for Computational Linguistics.

\bibitem[{Luong et~al.(2015)Luong, Pham, and
  Manning}]{luong-pham-manning:2015:EMNLP}
Minh-Thang Luong, Hieu Pham, and Christopher~D. Manning. 2015.
\newblock \href {http://aclweb.org/anthology/D15-1166} {Effective approaches to
  attention-based neural machine translation}.
\newblock In \emph{Proceedings of the Conference on Empirical Methods in
  Natural Language Processing}, pages 1412--1421. Association for Computational
  Linguistics.

\bibitem[{Neubig et~al.(2017)Neubig, Dyer, Goldberg, Matthews, Ammar,
  Anastasopoulos, Ballesteros, Chiang, Clothiaux, Cohn, Duh, Faruqui, Gan,
  Garrette, Ji, Kong, Kuncoro, Kumar, Malaviya, Michel, Oda, Richardson,
  Saphra, Swayamdipta, and Yin}]{dynet}
Graham Neubig, Chris Dyer, Yoav Goldberg, Austin Matthews, Waleed Ammar,
  Antonios Anastasopoulos, Miguel Ballesteros, David Chiang, Daniel Clothiaux,
  Trevor Cohn, Kevin Duh, Manaal Faruqui, Cynthia Gan, Dan Garrette, Yangfeng
  Ji, Lingpeng Kong, Adhiguna Kuncoro, Gaurav Kumar, Chaitanya Malaviya, Paul
  Michel, Yusuke Oda, Matthew Richardson, Naomi Saphra, Swabha Swayamdipta, and
  Pengcheng Yin. 2017.
\newblock Dynet: The dynamic neural network toolkit.
\newblock \emph{arXiv preprint arXiv:1701.03980}.

\bibitem[{Papineni et~al.(2002)Papineni, Roukos, Ward, and Zhu}]{Papineni:02}
Kishore Papineni, Salim Roukos, Todd Ward, and Wei-Jing Zhu. 2002.
\newblock \href {https://doi.org/10.3115/1073083.1073135} {{BLEU}: A method for
  automatic evaluation of machine translation}.
\newblock In \emph{Proceedings of the 40th Annual Meeting on Association for
  Computational Linguistics}, pages 311--318. Association for Computational
  Linguistics.

\bibitem[{Sennrich et~al.(2016)Sennrich, Haddow, and Birch}]{Sennrich:16}
Rico Sennrich, Barry Haddow, and Alexandra Birch. 2016.
\newblock \href {http://www.aclweb.org/anthology/P16-1162} {Neural machine
  translation of rare words with subword units}.
\newblock In \emph{Proceedings of the 54$^{th}$ Annual Meeting of the
  Association for Computational Linguistics}, pages 1715--1725.

\bibitem[{Smith et~al.(2013)Smith, Saint-Amand, Callison-Burch, Plamada, and
  Lopez}]{Smith:13}
Jason~R. Smith, Herve Saint-Amand, Chris Callison-Burch, Magdalena Plamada, and
  Adam Lopez. 2013.
\newblock \href {http://aclweb.org/anthology//P/P13/P13-1135.pdf} {Dirt cheap
  web-scale parallel text from the common crawl}.
\newblock In \emph{Proceedings of the Conference of the Association for
  Computational Linguistics}.

\bibitem[{Sukhbaatar et~al.(2015)Sukhbaatar, Szlam, Weston, and
  Fergus}]{Sukhbaatar:15}
Sainbayar Sukhbaatar, Arthur Szlam, Jason Weston, and Rob Fergus. 2015.
\newblock End-to-end memory networks.
\newblock In \emph{Proceedings of the 28th International Conference on Neural
  Information Processing Systems}, pages 2440--2448. MIT Press.

\bibitem[{Sutskever et~al.(2014)Sutskever, Vinyals, and Le}]{Sutskever:14}
Ilya Sutskever, Oriol Vinyals, and Quoc~V. Le. 2014.
\newblock Sequence to sequence learning with neural networks.
\newblock In \emph{Proceedings of the 27th International Conference on Neural
  Information Processing Systems}, pages 3104--3112. MIT Press.

\bibitem[{Wang et~al.(2017)Wang, Tu, Way, and Liu}]{Wang:17}
Longyue Wang, Zhaopeng Tu, Andy Way, and Qun Liu. 2017.
\newblock \href {http://aclweb.org/anthology/D17-1300} {Exploiting
  cross-sentence context for neural machine translation}.
\newblock In \emph{Proceedings of the Conference on Empirical Methods in
  Natural Language Processing}, pages 2816--2821. Association for Computational
  Linguistics.

\bibitem[{Weston et~al.(2015)Weston, Chopra, and Bordes}]{Weston:15}
Jason Weston, Sumit Chopra, and Antoine Bordes. 2015.
\newblock Memory networks.
\newblock In \emph{Proceedings of the International Conference on Learning
  Representations}.

\end{thebibliography}
\bibliographystyle{acl_natbib}


\end{document}